\pdfoutput=1
\documentclass[11pt]{article}
\usepackage{acl2023}
\usepackage{times}
\usepackage{latexsym}
\usepackage{multirow}
\usepackage[T1]{fontenc}
\usepackage[utf8]{inputenc}
\usepackage{microtype}
\usepackage{booktabs}
\usepackage{tabularx}
\usepackage{soul}
\usepackage{inconsolata}
\usepackage{graphicx}
\newcommand{\rt}[1]{\rotatebox{90}{#1}}
\usepackage{xspace}
\newcommand{\F}{$\textrm{F}_1$\xspace}
\usepackage{paralist}
\usepackage{bbding}

\title{Affective Natural Language Generation
  of\\ Event Descriptions through Fine-grained Appraisal Conditions}

\author{Yarik Menchaca Resendiz \and Roman Klinger\\
  Institut f\"ur Maschinelle Sprachverarbeitung, University of Stuttgart\\
  \texttt{\{yarik.menchaca-resendiz,roman.klinger\}@ims.uni-stuttgart.de}
}

\newcommand{\std}[1]{{\fontsize{6pt}{36pt}\selectfont({#1})}}
\newcommand{\countstd}[2]{\shortstack{{#1} \std{{#2}}}}

\begin{document}
\maketitle
\begin{abstract}
  Models for affective text generation have shown a remarkable
  progress, but they commonly rely only on basic emotion theories or
  valance/arousal values as conditions. This is appropriate when the
  goal is to create explicit emotion statements (``The kid is
  happy.''). Emotions are, however, commonly communicated
  implicitly. For instance, the emotional interpretation of an event
  (``Their dog died.'') does often not require an explicit emotion
  statement. In psychology, appraisal theories explain the link
  between a cognitive evaluation of an event and the potentially
  developed emotion. They put the assessment of the situation on the
  spot, for instance regarding the own control or the responsibility
  for what happens. We hypothesize and subsequently show that
  including appraisal variables as conditions in a generation
  framework comes with two advantages. (1)~The generation model is
  informed in greater detail about what makes a specific emotion and
  what properties it has. This leads to text
  generation that better fulfills the condition. (2) The variables of
  appraisal allow a user to perform a more fine-grained control of the
  generated text, by stating properties of a situation instead of only
  providing the emotion category.  Our Bart and T5-based experiments
  with 7 emotions (Anger, Disgust, Fear, Guilt, Joy, Sadness, Shame),
  and 7 appraisals (Attention, Responsibility, Control, Circumstance,
  Pleasantness, Effort, Certainty) show that (1) adding appraisals
  during training improves the accurateness of the generated texts by
  10\,pp in \F. Further, (2) the texts with appraisal
  variables are longer and contain more details. This exemplifies the greater
  control for users.
  
\end{abstract}

\section{Introduction}
\begin{figure}
  \centering\small
  \setlength{\tabcolsep}{1pt}
  \begin{tabular}{ll}\hline
    Condition: & {\setlength{\fboxsep}{0pt}\colorbox{blue!30!white}{Joy} \colorbox{green!30!white}{Responsibility}} \\
    Output: & {\setlength{\fboxsep}{0pt}\colorbox{blue!30!white}{I won
              the tournament} \colorbox{green!30!white}{due to
              extensive training}}.\\
    \hline
  \end{tabular}
  \caption{Conditioning text generation on emotions (blue) and appraisals (green) results in an improved fulfillment of the emotion condition by incorporating event descriptions (green) in the output text. This enables more fine-grained control over the generated text.}
  \label{fig:task-example}
\end{figure}

The main task of conditional natural language generation (CNLG) is to
provide freedom to control the output text. It is commonly addressed
as the intersection of text-to-text \citep{radford2019language,
  lewis2019bart,raffel2019exploring} and data-to-text generation
\citep{kondadadi2013statistical, lebret2016neural,
  ferreira2017linguistic}. Therefore, models typically use two inputs:
a textual trigger-phrase, and a condition to guide the generation.

In affective CNLG models, the condition is an affective state,
typically represented as valence/arousal values
\citep{maqsud2015synthetic} or discrete emotion names
\citep{ghosh-etal-2017-affect, song2019generating}. Arguably, the use
of theories of basic emotions \citep{ekman1994nature,
  plutchik2013theories} is appropriate when the main requirement is to
express a particular emotion. However, a natural communication of
emotions also includes implicit expressions, where the main content of
a message is not (only) the emotion. As an example, humans describe an
event and leave it to the dialogue partner to infer the affective
meaning (``Yesterday, my dog died''). In fact,
\citet{casel-etal-2021-emotion} report that event descriptions are
used to convey an emotion in 75\,\% of instances in the
TEC corpus \citep{Mohammad12}: The sentence ``I won money in the
lottery'' does, for most people, not require a mention of the
associated emotion.

In this paper, we focus on the task of generating such emotionally
connotated event descriptions (Figure \ref{fig:task-example}).
This poses the challenge how to represent the link
between ``factual'' events and their emotion. Appraisal theories from
psychology attempt to explain that connection with variables that
represent the cognitive evaluation by a person in context of a situation
\cite{Ellsworth1988,scherer2001appraisal}. Does the person feel
\textit{responsible}? Do they pay \textit{attention} to what is going on? Is the event \textit{pleasant}?
Does somebody have \textit{control} over what is happening? How much \textit{effort} is
needed to deal with the outcome of the situation? These variables
explain emotions: Feeling \textit{responsible} is a prerequisite for feeling
\textit{guilty}, not knowing about the outcome of a potentially negative event
might cause \textit{fear} (while knowing about it is more likely to cause
\textit{sadness}).

Our paper has two main contributions: (1) We hypothesize and show that
providing appraisal information along the emotion category to the
model, leads to a better fulfillment of the emotion condition. (2) We
show that adding appraisal variables leads to a more fine-grained
control of the generation process and the resulting texts show more
details regarding the described event.\footnote{Training scripts and 
generated data are available at \url{https://www.ims.uni-stuttgart.de/data/emotioncnlg}.}

\section{Related Work}
\subsection{Emotion and Appraisal Theories}
\label{sec:relatedworkemotion}
Emotions, a state of belief \citep{Green1992} that results in
psychological and physical changes, reflect individual's thoughts and
conduct. \citet{ekman1992argument} claims the existance of six basic
emotions (Anger, Disgust, Fear, Happiness, Sadness, and Surprise) that
occur in response to some stimulus. \citet{plutchik2001nature}
conceptualized eight primary emotions that serve as the foundation for
others. While these theories do mention events as a major element in
the process of developing an emotion, they do not explicitly explain
the link between stimulus events and the emotion category.

Appraisal theories aim at explaining the underlying cognitive process
of event evaluations. They link emotions via interpretations,
evaluations, and explanations of events.  \citet{Smith1985} show that
6 appraisal dimensions are sufficient to discriminate between 15
emotion categories---indeed, they constitute the
emotion. \citet{scherer2001appraisal} describes a sequence of
appraisals in which events are evaluated.

Appraisal theories have only recently received interest in
computational linguistics, firstly by developing analysis methods
motivated to analyze events and their structure \citep{Balahur2011}.
\citet{Hofmann2020b} were the first who explicitly modeled appraisal
variables in an existing corpus of event descriptions
\citep{troiano-etal-2019-crowdsourcing}. They used the variables from
\citet{Smith1985}, namely Attention, Certainty, Circumstance, Control,
Effort, Pleasantness, and Responsibility.  \citet{Troiano2023} created
a larger corpus and showed that appraisals can be reliably recovered
by external readers, and that they help for emotion classification. We
use their corpus
crowd-enVENT\footnote{\url{https://www.ims.uni-stuttgart.de
    /data/appraisalemotion}} of 6600 event descriptions, but limit
their (partially correlating) 21 appraisal concepts to those that
overlap with the definitions by \citet{Smith1985}, which were defined
via principle component analysis.

\subsection{Affective Natural Language Generation}
Most state-of-the-art systems for natural language generation follow a
sequence-to-sequence approach
\citep{sutskever2014sequence,cho-etal-2014-learning}. Such models take
as input a sequence of words and generate as output a sequence of
words. Chatbots, for instance, consider a question or an utterance
from the user as input and output an answer or reaction. The
architecture has two main modules, an encoder, which generates an
abstract semantic representation of the input text, and a decoder,
which takes the encoder representation and generates output words
\citep{sutskever2014sequence, radford2019language,
  raffel2019exploring, lewis2019bart}.

Transformer-based approaches commonly outperform recurrent neural
networks \cite{raffel2019exploring}. We use two such methods in our
paper, namely \textit{Bart} \citep{lewis2019bart}, which can be seen
as a generalization of GPT
\citep{radford2018improving,NEURIPS2020_1457c0d6,radford2019language}
for its left-to-right decoder and BERT \citep{devlin-etal-2019-bert}
due to the bidirectional encoder. The training objective is to
reconstruct the original text using a corrupted input. Further, we
use \textit{T5}, an encoder--decoder model with the philosophy to
reframe NLP problems as text-to-text tasks
\cite{raffel2019exploring}.

\begin{table}[t]
  \centering\small
\begin{tabularx}{\linewidth}{lX}
\toprule
  Conf. & Input Prompt and Output \\
  \cmidrule(r){1-1}\cmidrule(l){2-2}
  E & \ul{generate \textit{joy}:  \textit{Last day I}} \textbf{was very relaxed.} \\
EA & \ul{generate \textit{joy attention NoRESP control NoCIRC NoPLEA effort NoCERT}: \textit{Last
day I}} \textbf{was very relaxed because I worked for  6 hours} \\
A & \ul{generate \textit{attention NoRESP control NoCIRC NoPLEA effort NoCERT}: \textit{Last day
I}}
\textbf{decided to work for 6 hours} \\
  \bottomrule
\end{tabularx}
\caption{Examples for training data. The input prompt is  \ul{underlined},
conditions and trigger-phrase are in \textit{italic text}, and  the
output is printed in \textbf{bold}.}
\label{tab:eg-prompt}
\end{table}

Most conditional language generation work has focused on sentiment
polarity \citep{zhang2019emotional, maqsud2015synthetic,
  niu2018polite} and topical text generation
\cite{orbach2020facts2story,chan2020cocon}.  The small number of
papers that tackle emotion conditions include Affect-LM
\cite{ghosh-etal-2017-affect}, a language model for generating
conversational text, conditioned on five categories (Anger, Sadness,
Anxiety, Positive, and Negative sentiment). Affect-LM enables
customization of emotional content and intensity in the generated
sentences. The customization is achieved by concatenating a condition
vector to the embedding representation of the sentence. EmoDS
\citep{song2019generating} is a dialogue system that can generate
responses expressing the desired emotion explicitly or implicitly. The
implicit generation is guided by a sequence-level emotion classifier,
which recognizes a response not containing any emotion word. Within
the dialog domain, the Emotional Chatting Machine involves three
modules to generate responses \citep{zhou2018emotional}.  These
modules are a high-level abstraction of emotion expressions, a change
in implicit internal emotion states, and an external emotion
vocabulary. The Multi-turn Emotional Conversation Model
\cite[MECM,][]{cui2022modeling} introduces modules to track the
emotion throughout the conversation.  \citet{colombo-etal-2019-affect}
presents a GPT-2-based model \citep{radford2019language}. They use
classifiers together with emotion and topic lexicons to guide the 
output. We use this model as a strong baseline.

None of the previous works focused on generating emotionally
connotated event descriptions, which are a natural way to tell someone
about the own emotional experience. None of them used
psychological theories other than affect and basic emotions. We fill
these gaps by combining the recent methods with appraisal theories.

\section{Methods}
\label{sec:Methods}
The objective of our paper is to understand if adding appraisal
information in addition to emotion conditions to a generator (1)
improves the accuracy of the output, i.e., the likelihood that the
output in fact exhibits the target emotion. Further, (2), we aim at
understanding if these appraisal variables provide a more fine-grained
control to the users (e.g., ``I am relaxed'' vs.\ ``I am relaxed
because I worked for only 6 hours'').  To address these goals, we
configure three CNLG models (Table \ref{tab:eg-prompt}), all based on
\textit{Bart} \citep{lewis2019bart} and \textit{T5}
\cite{raffel2019exploring}: (a) \textit{Condition on emotions} (E),
where the model only gets informed by the target emotion (Anger,
Disgust, Fear, Guilt, Joy, Sadness, or Shame) to be expressed in the
generated text. (b) \textit{Condition on emotions and appraisals}
(EA), which has both the emotions and appraisals as conditional
variables. The comparison between E and EA will allow us to understand
the impact of the appraisals. In addition (c), we \textit{condition on
  appraisals} only (A), where each generated sentence can be
conditioned on one or multiple appraisals (Attention, Responsibility,
Control, Circumstance, Pleasantness, Effort, or/and Certainty).

\paragraph{Training.} In each configuration, we embed the conditions in the input prompt, to 
fine-tune the models. This strategy avoids expensive training---encoders
or decoders, or both---with condition information from scratch. We
create training data out of existing corpora that are labeled for
emotions and appraisals consisting of input prompts and output pairs.
The input prompt contains the conditions (e.g., \textit{joy}; \textit{joy attention}),
as special tokens, followed by the trigger-phrase (e.g.,
\textit{Last day I}). The output are the remaining words the model should learn to produce
(e.g., was relaxed because I worked for 6 hours). 
This leads to the following three prompt representations (see
Table~\ref{tab:eg-prompt} for examples):
\begin{compactdesc}
\item [E:] (condition on emotions only)\\ 
  ``generate  [\textit{emotion}]: [\textit{trigger-phrase}]''
\item [EA:] (condition on both emotions and appraisals)\\ 
  ``generate  [\textit{emotion}] [\textit{appraisals}]$^m$: [\textit{trigger-phrase}]''
\item [A:] (condition on appraisals only)\\ 
  ``generate  [\textit{appraisals}]$^m$: [\textit{trigger-phrase}]''
\end{compactdesc}
where $\textit{emotion}$ $\in$ \{anger, shame, disgust, fear, guild,
joy, sadness\} and \textit{appraisals} is a string of the form
``\{attention, NoATTE\} \{responsibility, NoRESP\} \{control, NoCONT\}
\{circumstance, NoCIRC\} \{pleasantness, NoPLEA\} \{effort, NoEFFORT\}
\{certainty, NoCERT\}''. The \textit{trigger-phrase} consists of the first $n$
words of the training text, where $n$ is randomly chosen
($1\leq n \leq 9$).

By using non-special tokens to represent the target conditions, the
models can make use of knowledge acquired in pretraining. We opt for a
string representation over a numerical representation (e.g.,
``control'' instead of ``1'' or ``NoCONT'' instead of ``0''), because
preliminary experiments showed that numerical representations are
sometimes interpreted as a request for repetitions by \textit{T5}
(\textit{``generate 1 1: I feel''} $\rightarrow$ \textit{``I feel I
  feel''}).

\paragraph{Inference.} At prediction time, we obtain the five most
probable sentences for each prompt. These sentences are selected using
beam search \citep{lowerre1976harpy} with beam size 30, next token
temperature of 0.7, top-p\footnote{Top tokens whose sum of likelihoods
  does not exceed a certain value (p).}  (nucleus) sample of 0.7. We
ensure that our output excludes sentences with repeated
instances of the same bigram.

\section{Experiments}
\label{sec:experiments}
The following subsections explain the experiments 
conducted to test our hypotheses. In \S\ref{sub:experiments}, we
describe the setting and fine-tuning of the models.  In
\S\ref{sec:Results}, we provide results to answer the question (1) if
appraisals in conjunction with emotion conditions improve the
generation such that it meets the emotion condition. In
\S\ref{sec:Qua-Analysis} we discuss results to understand if
appraisals are a means for a more fine-grained control of the
generation process.

\begin{figure}
\centering
  \includegraphics[scale=0.60]{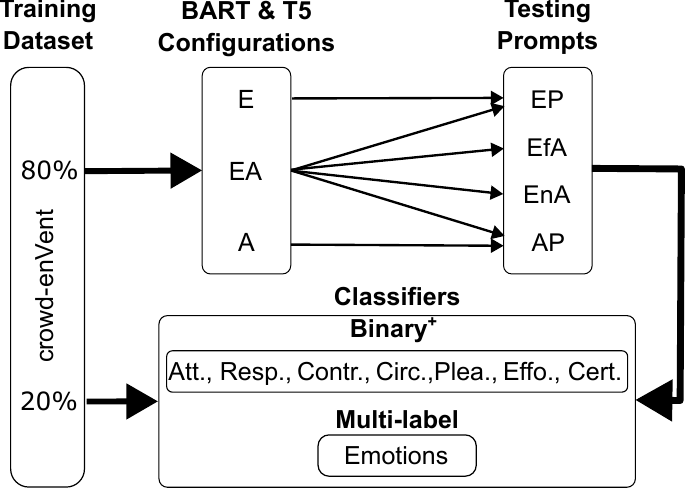}
  \caption{Experiment workflow}
  \label{fig:work-flow}
\end{figure}

\subsection{Experimental Settings}
\label{sub:experiments}
Figure~\ref{fig:work-flow} illustrates the workflow and the utilized
combinations between classifiers, CNLG models, and synthetic testing
prompt sets. We fine-tune according to three training set
configurations (E, EA, A). This leads to six models (Bart, T5) which
we evaluate with multiple testing prompt sets. The testing prompt sets
only partially mirror the training regime, because the combinations of
the conditional variables can be expected to be put together more
freely at prediction time than as they occur in labeled data. We
compare the emotion-informed models (E, EA) using the emotion testing
prompt set (EP) to understand the impact of adding appraisals in the
condition while not showing appraisals at prediction time. This
enables us to understand if presenting appraisals improves the model's
internal representation of emotion concepts.

In addition, to understand how appraisals influence the output at
inference time, we use testing prompt set with the most frequently
cooccurring appraisals (EfA)---these combinations can be considered to
be ``compatible'' with each other and the emotion (Figure
\ref{fig:work-flow}). To challenge the models, we further use the
emotion with all appraisals turned off (emotion with negative
appraisals, EnAP) and test what happens when we do not provide an
emotion category (appraisal-only, AP). To evaluate the performance of
the models, we calculate \F with automatic emotion and appraisal
classifiers (\S\ref{sec:experiments}) and with human annotation (\S
\ref{sec:Human-eva}).

\paragraph{Dataset.}
The basis for our experiments is the crowd-enVENT data set of
autobiographical reports of emotional events (see
\S\ref{sec:relatedworkemotion}).  We use a subset to train
emotion and appraisal classifiers for evaluation and another subset for
fine-tuning the generators (Appendix \ref{sec:filtered_enVent}). Each event
has 21 author-assessed appraisal values, created by asking
crowdworkers to complete a sentence for a given emotion (e.g., ``I
felt [emotion] when/that/if...''). We observed in preliminary
experiments that both generation architectures (\textit{T5} and
\textit{Bart}) have issues differentiating between the conditions and
the trigger phrase, potentially due to the incompatibility of the
conditions. For that reason, we focus on emotions and appraisals that
have been proven to be predictable by \citet{Hofmann2020b}---the
variables that \citet{Smith1985} showed to be principle components for
emotion categories.

We use instances that correspond to one of seven emotions (Anger,
Disgust, Fear, Guilt, Joy, Sadness, and Shame) and contain an
annotation with at least one of the seven appraisals\footnote{We
  discretize the [1:5] ordinal values to boolean values at a threshold
  of $\geq 4$, as suggested by the authors of the data set.}
(Attention, Responsibility, Control, Circumstance, Pleasantness,
Effort, and Certainty). This leads to 2750 instances in the corpus
that we use for training. Appendix \ref{sec:filtered_enVent} reports
details and statistics of our filtered data.

\paragraph{Model Training and Data Augmentation.}
We train the generation models with 80\,\% of the instances from this
filtered corpus. The dataset is preprocessed with two goals, firstly,
to create the prompts (\S\ref{sec:Methods}) according to the
desired model configuration (A, E, EA), and secondly to augment the
data to prevent the models from mapping the same trigger phrase to the
same output. To achieve that, we duplicate each instance $t$ times,
where $2\leq t \leq 5$ is randomly chosen. In each duplication, a
unique random number of $n$ token combinations ($1\leq n \leq 9$) from
the textual instance is used as part of the trigger phrase. Therefore, the
duplication does not lead to identical instances.

\paragraph{Emotion and Appraisal Classifiers.}
To evaluate the performance of the generation models automatically, we
use eight classifiers (one per appraisal and one for all emotions)
using the remaining 20\,\% of the filtered crowd-enVENT dataset
(15\,\% for training the classifier, and 5\,\% to evaluate the
classifiers). The classifiers are built on top of RoBERTa
\citep{liu2019roberta} with default parameters (10 epochs, batch size
5). Each appraisal classifier predicts a boolean value whereas the
emotion classifier predicts one of seven emotions. The classifiers
show a performance of .75 \F Macro-Avg.\ for emotion classification
and .56 \F for appraisal classification. These scores are, despite the
limited amount of available data, comparable to previous experiments
\cite{Troiano2023}. Details on these classifiers are reported in
Appendix \ref{sec:Auto Classifiers}. These classifiers allow us to
perform a large set of experiments, but the non-perfect performance
motivates us to confirm the main results in a human study
(\S\ref{sec:Human-eva}).

\paragraph{Evaluation.}
To evaluate the three CNLG model configurations, we create four
testing prompt sets each using the thirteen most frequent starting
n-grams from the crowd-enVent dataset (``I felt'', ``When a'', ``I
was'', ``When I'', ``I had'', ``I got'', ``When my'', ``I found'', ``I
went'', ``I saw'', ``I did'', ``When someone'', and ``I am'') as
trigger phrase, the seven emotions and the seven
appraisals. \textit{Emotion Prompt set} (EP) consists of 91 possible
combinations between prompts and emotions (e.g., \textit{generate joy:
  I felt}). The \textit{Emotion with most frequent Appraisals Prompt
  set} (EfA) includes the 910 combinations between prompts, emotions
and the 10 most frequent appraisals per emotion from the crowd-enVent
corpus. The \textit{Emotion with negative Appraisals Prompt sets}
(EnAP) is similar to EP, but includes the appraisal vector, all set to
negative values. The \textit{Appraisal Prompt set} (AP) has the 104
possible combinations between the 13 prompts and one appraisal at a
time (including the case where all appraisals are off).

It is nonsensical to compare all CNLG models on all testing prompt sets
(Figure \ref{fig:work-flow}, interaction between Bart \& T5 configurations
and Test Prompts)---e.g., the E configuration would not be
able to interpret appraisal prompts (AP), similarly for the A model
configuration. For every possible combination between CNLG model and
the four testing prompt sets, we generate the five most probable sentences for
each prompt (13,910 in total).

\paragraph{State-of-the-art Baseline.}
To understand how well a generic model can solve the task of affective
event generation, we compare against the Affective Text Generation
model \cite[ATG,][]{colombo-etal-2019-affect}. ATG is conditioned on
both an emotion and a topic, with the help of word lexicons. To make a
fair comparison with \textit{T5} and \textit{Bart}, we fine-tune the language model
underlying ATG, namely GPT-2, to produce emotion event descriptions
using the same data that we use to train \textit{T5} and
\textit{Bart}.  The emotion and topic lexicons are unmodified
because we consider them to be an essential element of ATG. Finally,
for each emotion that is available in ATG and in our data (Fear, Joy,
Anger, Disgust, Sadness), we generate sentences with varying intensity
and target topic (Legal, Military, Politics, Monsters,
Religion, Science, Space, Technology---520 in total).

\subsection{RQ1: Do Appraisal Variables Improve Affective Text Generation?}
\label{sec:Results}
We start the discussion of our first goal of this paper (do appraisal
variables improve the model) quantitatively. Table~\ref{tab:cnlg-emotions}
shows how well the texts
from the various generation models exhibit the target emotion
(evaluated against the automatic classifiers). The results should be
interpreted in the context of the perplexity (Ppl.) information in
Table~\ref{tab:auto-stats}.

\begin{table}
  \centering\small
  \newcommand{\sep}{\cmidrule(r){1-1}\cmidrule(r){2-2}\cmidrule(){3-3}\cmidrule(rl){4-4}\cmidrule(rl){5-5}\cmidrule(rl){6-6}\cmidrule(rl){7-7}\cmidrule(rl){8-8}\cmidrule(rl){9-9}\cmidrule(l){10-10}\cmidrule(l){11-11}}
  \setlength{\tabcolsep}{3pt}
  \renewcommand{\arraystretch}{0.95}
  \begin{tabular}{lllrrrrrrrr}
    \toprule
    \rt{Arch.} & \rt{Conf.} &  \rt{\shortstack{Testing\\ Prmpt.}} & \rt{Ang.} & \rt{Disg.} & \rt{Fear}& \rt{Guilt} & \rt{Joy} & \rt{Sad.} & \rt{Shame} & \rt{M. Avg.} \\
    \sep
    ATG & E & --- & .10 & .18 & 0.25 & --- & .06 & .17 & --- & .15 \\
    \sep
    T5 & E  & EP & .28 & .50 & .63 & .23 & .60 & .32 & \textbf{.40} & .42\\
    T5 & EA & EP & .46 & \textbf{.58} & \textbf{.70} & .27 & \textbf{.77}&\textbf{.58} & .32 & \textbf{.52}\\
    T5 & EA & EfA & .39 & .60 & .57 & \textbf{.35} & \textbf{.77} & .47 & .21 &.48\\
    T5 & EA & EnAP & \textbf{.52} & .55 & .64 & \textbf{.35} & .58 & .41 & .19 &.46\\
    \sep
    Bart & E & EP & .36 & .45 & .40 & .29 & .63 & .43 & \textbf{.49} & .43\\
    Bart & EA & EP & \textbf{.41} & \textbf{.57} & .48 & \textbf{.41} & .63 &\textbf{.54} & .36  & \textbf{.49}\\
    Bart & EA & EfA & .34 & .45 & \textbf{.52} & .29 & \textbf{.75} & .46 & .44 &.47\\
    Bart & EA & EnAP & .34 & .51 & .43 & .26 & .57 & .33 & .37 & .40\\
\bottomrule
\end{tabular}
\caption{Emotion F$_1$ scores of models trained with only emotions (E), emotions and
  appraisal conditions (EA), and only
  appraisal conditions (A) over the generated text using the testing prompt
  sets: EP (Emotions Prompt set), EnAP (Emotions with negative
  Appraisals Prompt set, all the appraisals are turned off) and EfA
  (Emotion with the most frequent Appraisals Prompt set).}
\label{tab:cnlg-emotions}
\end{table}

Table~\ref{tab:cnlg-emotions} confirms our hypothesis for both \textit{T5} (2nd
block) and \textit{Bart} (3rd block). The important parts are the E
and EA models compared on the same \textit{testing emotion prompt set} (EP), which only
contains emotion conditions. We see that, except for Shame, the
appraisal-informed model always shows a better performance---despite
not showing appraisal information at inference time. Apparently, the
model learns a more accurate internal emotion representation with the
additional information. On average, \textit{T5} shows a 10pp higher \F
with appraisal information than without.

Obviously, an interesting question is if this performance could be
further improved when providing additional appraisal information to
the prompt. When using appraisal values frequently cooccurring with
the emotion concept (EfA), the performance is still higher than when
not providing appraisal values during training, but apparently leaving
the model more freedom in the generation with fewer conditions leads
to better texts (EfA vs.\ E). As expected, turn off the appraisals
(EnAP) leads to a drop in performance---but remains still better than
the emotion-only (E) models.

Across all experiments, \textit{T5} outperforms \textit{Bart} and
ATG. The low ATG performance could be attributed to the use of
dictionaries to guide the generation process, which naturally has
limited coverage and might not be suitable to describe events.

These results need to be interpreted in context with the perplexity
scores shown in the last column of Table~\ref{tab:auto-stats}. Here,
we see that ATG shows better performance. More importantly to answer
our research question regarding the impact of appraisals is to compare
the perplexity of the various E, EA, and A configurations. For the \textit{T5}
model (which shows the better emotion accuracy), there is a small
decrease in language quality measured with perplexity. For the Bart
model, the perplexity is in fact improving with appraisals.

\begin{table}[t]
  \newcommand{\sepp}{\cmidrule(r){1-1}\cmidrule(rl){2-2}\cmidrule(rl){3-3}\cmidrule(rl){4-4}\cmidrule(rl){5-5}\cmidrule(rl){6-6}\cmidrule(rl){7-7}\cmidrule(l){8-8}}
  \centering\small
  \setlength{\tabcolsep}{1.5pt}
  \begin{tabular}{lllrrrrrr}
    \toprule
    \rt{Arch.} & \rt{Conf.} &  \rt{\shortstack{Testing\\ Prmpt.}}& \rt{\shortstack{Tokens\\ (std.)}} &\rt{\shortstack{Nouns\\ (std.)}} &\rt{\shortstack{Verbs\\ (std.)}} &\rt{\shortstack{Clauses\\ (std.)}} & \rt{Ppl.} \\
    \sepp
    Hum. & Hum. & enVent & \countstd{19.3}{23} & \countstd{3.2}{3.5} &\countstd{2.8}{3.3} & \countstd{.9}{1.5}  & --- \\
    \sepp
    ATG & E & --- & \countstd{16.4}{1.6} & \countstd{2.4}{1.3} & \countstd{2.3}{.9} & \countstd{1.7}{.6} & 22.2 \\
    \sepp
    T5 & E & EP & \countstd{9.2}{3.4} & \countstd{2.1}{1.0} & \countstd{2.2}{1.0}& \countstd{1.2}{.6} & 26.9\\
    T5 & EA & EP & \countstd{15.1}{4.3} & \countstd{2.3}{1.1} &\countstd{2.3}{1.1} & \countstd{1.5}{.6} &  28.5\\
    T5 & EA & EfA & \countstd{13.9}{4.8} & \countstd{2.1}{1.1} &\countstd{2.1}{1.1} & \countstd{1.5}{.6} & 28.5\\
    T5 & EA & EnAP & \countstd{14.3}{4.5} & \countstd{2.2}{1.0} &\countstd{2.2}{1.1} & \countstd{1.5}{.6} &  28.5\\
    T5 & A & AP & \countstd{8.2}{3.8} & \countstd{1.8}{1.1} & \countstd{1.8}{1.0} & \countstd{1.2}{.6} & 23.5\\
    \sepp
    Bart & E & EP & \countstd{8.1}{4.1} & \countstd{1.7}{1.1} & \countstd{1.9}{1} &\countstd{1.4}{.5} & 69.2\\
    Bart & EA & EP & \countstd{10.5}{3.7} & \countstd{1.9}{1.0} &\countstd{1.6}{.8} & \countstd{1.2}{.4} & 51.3\\
    Bart & EA & EfA & \countstd{11.7}{4.1} & \countstd{1.9}{1.1} &\countstd{1.8}{1} & \countstd{1.3}{.5} & 51.3\\
    Bart & EA & EnAP & \countstd{13.2}{4.4} & \countstd{2.3}{1.1} &\countstd{1.9}{1} & \countstd{1.4}{.6} & 51.3\\
    Bart & A & AP & \countstd{7.7}{3.4} & \countstd{1.7}{1.2} &\countstd{1.4}{1.2} & \countstd{1.4}{.4} & 58.3\\
\bottomrule
\end{tabular}
\caption{Analysis of generated text using different
  model architectures, configurations, and prompt test
  sets. Mean/standard deviations are based on Spacy's tokenizer and
  POS. Ppl.: perplexity on test data.}
\label{tab:auto-stats}
\end{table}

\begin{table*}[t]
  \centering\small
  \setlength{\tabcolsep}{2pt}
\renewcommand{\arraystretch}{1}
\begin{tabularx}{\linewidth}{llp{30mm}lX}
\toprule
N. & Conf. & Condition & Trigger-Phrase & Generated Text \\ 
\cmidrule(r){1-1}\cmidrule(rl){2-2}\cmidrule(rl){3-3}\cmidrule(rl){4-4
}\cmidrule(l){5-5}
1 & E & Shame & I felt & I felt ... when I was in my early 20's. \\
2 & EA & Shame Att., Resp., Effo. & I felt & I felt ... when I walked out of a
class meeting with no explanation \\
3 & EA & Shame and Resp. & I felt & I felt ... when I walked out of a
restaurant and smelt bad \\
\cmidrule(r){1-1}\cmidrule(rl){2-2}\cmidrule(rl){3-3}\cmidrule(rl){4-4
}\cmidrule(l){5-5}
  4 & E & Joy & I got & I got a job I really wanted. \\
5 & EA & Joy,  Resp., Contr., Plea. & I got & I got a job I had wanted for
months leading up to my graduation.\\
6 & EA & Joy, Resp., Contr., Plea. Effo. & I got & I got a job I applied for
last week. \\
\cmidrule(r){1-1}\cmidrule(rl){2-2}\cmidrule(rl){3-3}\cmidrule(rl){4-4
}\cmidrule(l){5-5}
  7 & E & Fear & I had & I had to go on a rollercoaster \\
8 & EA & Fear, Att. & I had & I had to go into a hospital with covid symptoms
and give birth \\
9 & EA & Fear, Att., Effo., Cert. & I had & I had to go into a hospital because
I didn't know what was wrong with me. \\
\cmidrule(r){1-1}\cmidrule(rl){2-2}\cmidrule(rl){3-3}\cmidrule(rl){4-4
}\cmidrule(l){5-5}
  10 & E & Sad. & When my & When my dad passed away \\
11 & EA & Sad., Effo., Cert. & When my & when my boyfriend told me he was
leaving for college because I was too busy to do anything with him\\
12 & EA & Sad., Plea. & When my & when my boyfriend told me he was leaving
for college because I felt alone and I like I was going too \\
\cmidrule(r){1-1}\cmidrule(rl){2-2}\cmidrule(rl){3-3}\cmidrule(rl){4-4
}\cmidrule(l){5-5}
  13 & E & Ang. & I found & I found out my partner was cheating on me \\
14 & EA & Ang.,  Plea. & I found & I found out my partner was cheating on me
\\
15 & EA & Anger, Effo., Cert. & I found & I found out my partner was taking
advantage of my offer to buy him a house.\\
\cmidrule(r){1-1}\cmidrule(rl){2-2}\cmidrule(rl){3-3}\cmidrule(rl){4-4
}\cmidrule(l){5-5}
  16 & E & Disg. & I went & I went to the doctor and found a dead frog. \\
17 & EA & Disg., Att., Effo. & I went & I went to a restaurant to try their
dishes. They were very poor quality and did not clean up. \\
18 & EA & Disg., Effo. & I went & I went to a restaurant to try
                                    their dishes.' \\
\cmidrule(r){1-1}\cmidrule(rl){2-2}\cmidrule(rl){3-3}\cmidrule(rl){4-4
}\cmidrule(l){5-5}
19 & E & Guilt & I saw & I saw a friend being bullied at school. \\
20 & EA & Guilt, Resp. & I saw & I saw a homeless person who needed medical
attention because I couldn't  afford it \\
21 & EA & Guilt, Resp., Cert. & I saw & I saw a homeless person who had been
ill and died \\ \bottomrule
\end{tabularx}
\caption{Example texts generated by T5 using different model configurations,
conditions, and Trigger-Phrases.}
\label{tab:examples}
\end{table*}

\subsection{RQ2: Do Appraisals Allow for a more Fine-grained Control?}
\label{sec:Qua-Analysis}
To understand how appraisal theories can provide a more fine-grained
control to the user, we conduct a quantitative and a qualitative
analysis.

\paragraph{Quantitative Analysis.} Table~\ref{tab:auto-stats} shows
the statistics of the generated data with the various model
configurations for various prompts and as a point of reference the
human and ATG-model results. Under the assumption that appraisals
provide more information and more control, we would expect longer,
more detailed instances with the EA models. This is indeed the case
for both \textit{T5} and \textit{Bart}. On the emotion prompt test set
(EP), instances obtained with the model trained with appraisal
information (EA) are 15 tokens long for \textit{T5}, while instances
of the model trained only with emotion conditions (E) are 9 tokens
long. When adding incompatible appraisal information to the prompt
test data (EnAP), the text becomes even longer, with 15 tokens. The
compatible appraisal values (EfA) are in between with 14 tokens.  The
perplexity is mainly influenced by the model architecture (GPT-2 being
best, closely followed by \textit{T5}), but it is lower for
appraisal-informed models. Therefore, we can conclude that EA models
generate longer instances, however, it is accompanied by the drawback
of text quality, as evidenced by an increase in perplexity.

\paragraph{Qualitative Analysis.}
To gain a better understanding of the impact of appraisal information
on the generated text, we focus on \textit{T5}, the best-performing
model to generate the target emotion
(\S\ref{sec:Qua-Analysis}). Table~\ref{tab:examples} shows examples of
texts that stem from different configurations (same trigger-phrase but
different conditions). We select the most frequent appraisal and
emotion combinations from the crowd-enVent dataset as conditions to
generate texts.

We see that E-configuration-based generation lacks details on the
event in comparison to the EA configuration (Sentence 4 vs.\ 5 or
6). In Sentence 5, ``I had wanted for months leading up to my
graduation.'' the graduation aspect of the event makes one's
responsibility for getting a desired job more prominent. Such
properties can similarly be found in other sentence pairs in the E
(e.g., 1, 4, 7, 13, 16) and EA (e.g., 2, 3, 5, 15, 17) configurations.

Appraisals that are untypical for an emotion (e.g., \textit{Pleasantness} in
\textit{Fear} or \textit{Sadness}) do not change the general emotion of the text (e.g., 13
and 14), but they guide the models in order to describe an event that
fulfills the appraisal condition. This can be seen in a comparison of
Sentences 11 and 12, where the difference is a switch of \textit{Certainty} and \textit{Effort}
to \textit{Pleasantness}. The model then generates ``I like I was going...'' to
add some pleasantness despite the predominant condition being
\textit{Sadness}. Other cases show that the appraisal
condition is ignored by the generator if the emotion condition is
contradicting (Sentence 13 and 14). This explains why
 EnAP testing prompts show longer results (Table~\ref{tab:auto-stats}).

\section{Human Evaluation}
\label{sec:Human-eva}

We conduct a human study to validate the automatic evaluation.
Further, this study assesses additional measures,
namely the quality of the generated text. We focus on the
best-performing model, \textit{T5}, fine-tuned in the EA and E configurations.

\paragraph{Setup.}  We randomly select 100 sentences from the
following model-configuration and testing prompt set combinations: EA
with EP, E with EP, and EA with EfA. In addition, we include 30
sentences from the crowd-enVent dataset to confirm the validity of the
crowd-working setup. These 30 sentences are selected to be
``easily-annotated'' based on a high inter-annotator agreement in the
original data.

We evaluate the 330 sentences on the platform
\url{https://www.soscisurvey.de}. The survey consists of 23 statements
to be rated on a five-level Likert scale. Seven statements correspond
to the emotions (``What do you think the writer of the text felt when
experiencing this event?''). Seven statements correspond to the
appraisal variables (``How much do these statements apply?''), and
seven questions measure the text quality (fluency, grammaticality,
being written by a native speaker, semantical coherence, realistic
event, written by an artificial intelligence, written by a human). In
addition, we include two attention checks. We recruit participants via
\url{https://www.prolific.co/}.  \S \ref{sec:Human_suervey} shows the
questions in detail.

\paragraph{Results.} To compare the performance of the conditional natural language generation models,
using the human evaluation (five-level), we discretize emotion and
appraisal scores, analogously to the discretization of the
crowd-enVENT labels for our conditional models.  We assign the labels
based on a majority vote of three annotators.

\begin{table}[t]
  \newcommand{\seppp}{\cmidrule(r){2-2}\cmidrule(rl){3-3}\cmidrule(rl){4-4}\cmidrule(rl){5-5}\cmidrule(rl){6-6}\cmidrule(rl){7-7}\cmidrule(rl){8-8}\cmidrule(rl){9-9}\cmidrule(r){10-10}\cmidrule(rl){11-11}}
  \centering\small
\setlength{\tabcolsep}{3.5pt}
\begin{tabular}{lllrrrrrrrr}
\toprule
& \rt{Conf.} &  \rt{\shortstack{Testing\\ Prmpt.}} & \rt{Ang.} & \rt{Disg.} & \rt{Fear}& \rt{Guilt} & \rt{Joy} & \rt{Sad.} & \rt{Shame} & \rt{M. Avg.} \\
\seppp
\multirow{4}{*}{\rt{Hum.}}
& Hum. & enVent & 1 & 1 & 1 & \textbf{1} & 1 & 1 & 1& 1\\
\seppp
& E & EP & .69 & .72 & .72 & \textbf{.83} & .89 & .67& \textbf{.82} &\textbf{.76}\\
& EA & EP & \textbf{.79} & \textbf{.74} & \textbf{.73} & .62 & \textbf{.92} &\textbf{.82} & .6 & .74\\
& EA & EfA & .73 & .67 & .62 & .45 & .71 & .74 & .65 & .65\\
\cmidrule(rl){1-11}
\multirow{4}{*}{\rt{Auto.}}& Hum. & enVent & .86 & 1 & .9 & 1 & 1 & 1 & 1 & .97\\
\seppp
& E & EP & .46 & .14 & .0.5 & \textbf{.44} & .78 & .33 & \textbf{.41}& .44\\
& EA & EP & \textbf{.55} & \textbf{.38} & \textbf{.82} & .31 &\textbf{1} &\textbf{.6}& .26 & \textbf{.56}\\
& EA & EfA & .53 & .5 & .33 & .4 & .67 & .5 & .2 & .45\\
\bottomrule
\end{tabular}
\caption{Human annotation results as \F (top). For comparison,
  we show the automatic evaluation on the same subsample (bottom).}
\label{tab:human-emotions}
\end{table}

Table \ref{tab:human-emotions} shows the performance of the generation
models evaluated by the annotators on the top (Hum.). To be able to compare this
to the automatic evaluation that we reported in
\S\ref{sec:Results} we show the automatic classifier-based
evaluation on the same data that we used for human evaluation in
addition at the bottom (Auto.). The first row, in both the human and the automatic
evaluation, is the result of the evaluation on the 30 ``easily-annotated''
instances from the crowd-enVent data---both parts
perform close-to-perfect---confirming that the general experimental
setup is feasible. Further, we see that the automatic evaluation on
the subset used for human evaluation mimics the results in
Table~\ref{tab:cnlg-emotions}.

The two rows for the EP testing prompt (with EA and E model configurations)
also mimic the automatic evaluation. This is, however, not shown in
the average \F score because the differences are less
pronounced. Nevertheless, we observe that all emotions are better
generated with the EA model than with the E model, except for \textit{Guilt}
and \textit{Shame}. Therefore, the human evaluation confirms that training
models with appraisal information lead to a better generation of
emotion-bearing sentences. We report results for appraisals in
Appendix ~\ref{sec:Human_eval_Appraisals}.

Table \ref{tab:human-grammar} shows the results for the evaluation of
the quality of the generated sentences, in terms of fluency, grammar
errors, coherency, text origin (text was written by a native English
speaker or machine), and mimicking real event descriptions (what the
text describes might happen).
We have seen in Table~\ref{tab:auto-stats} that instances generated
with appraisal conditions in addition to emotion conditions lead to
considerably longer texts. This seems to come with the disadvantage
that the text quality is lower in all measured
variables. Nevertheless, most of the values are still in an acceptable
range, with the exception for grammaticality and the estimate that the
text might have been written by an AI (which, however, both show
comparably low values for real texts as well). As expected, the
variables \textit{Written by AI} and \textit{Written by Human} have a
strong negative correlation (Pearson's $\rho=-.77$).  Importantly, the
text mostly remains coherent.

\begin{table}
  \centering\small
\setlength{\tabcolsep}{3pt}
\begin{tabular}{llrrrrrrr}
\toprule
\rt{Conf.} &  \rt{\shortstack{Testing\\ Prmpt.}} & \rt{Fluency} & \rt{Grammar} &
\rt{\shortstack{Native\\ Spkr}}
& \rt{Coherency} & \rt{\shortstack{Really \\happen}} &
\rt{\shortstack{5$-$Written
\\by AI}} &
\rt{\shortstack{Written\\
by Human}}
\\
\cmidrule(r){1-1}\cmidrule(rl){2-2}\cmidrule(rl){3-3}\cmidrule(rl){4-4
}\cmidrule(rl){5-5}\cmidrule(rl){6-6}\cmidrule(rl){7-7}\cmidrule(rl){8-8
}\cmidrule(rl){9-9}
Hum. & enVent & 4.1 & 2.98 & 4 & 3.83 & 4.47 & 2.83 & 3.92\\
\cmidrule(r){1-1}\cmidrule(rl){2-2}\cmidrule(rl){3-3}\cmidrule(rl){4-4
}\cmidrule(rl){5-5}\cmidrule(rl){6-6}\cmidrule(rl){7-7}\cmidrule(rl){8-8
}\cmidrule(rl){9-9}
E & EP & \textbf{3.55} & \textbf{2.43} & \textbf{3.4} & \textbf{3.36}&
\textbf{4} &\textbf{2.42} &\textbf{3.25}\\
EA & EP & 3.07 & 1.88 & 2.82 & 2.89 & 3.57 & 1.86 & 2.93 \\
EA & EfA & \textbf{3.55} & \textbf{2.43} & 3.3 & 3.23 & 3.88 & 2.17 & 3.18\\
\bottomrule
\end{tabular}
\caption{Human evaluation of text quality using the five-level Likert scale,
where 1 is \textit{not agree at all}, and 5 is \textit{extremely
  agree}. (higher is better).}
\label{tab:human-grammar}
\end{table}

\section{Conclusion and Future Work}
We presented the first study on conditional text generation based on
both basic emotion category names and appraisal theories. We find that
the emotion is more reliably represented when appraisals are provided
during training, even when the appraisals are not provided during
inference.  

In addition, we provide evidence that the combination of
appraisals enables a more fine-grained control over the generated
text. By switching the appraisal variables, distinct event
descriptions are produced, even when the emotion remains constant.

This leads to important future work: While we believe that appraisals
shall be used to generate more detailed and accurate texts, the
decrease in text quality needs to be controlled. In our work, we
relied on prompt-based representations of the conditions in the
generator models. Different model architectures (e.g., embedding the
condition into the encoder, decoder, or both) could improve or
maintain the quality of the generated text.

In our experiments, we relied on annotated data with labels that we
used as conditions. In these data, all variables were always
accessible.  In a real-world setup, a deployable model would need to
automatically estimate (a subset of) appraisal dimensions or request
required information from a user. This might lead to a novel setup of
conditioning under partial information which poses new challenges for
general models of conditional text generation.

Finally, we left the topic of the event description to the choice of
the model. In a real-world setup, additional conditions need to be
included, for instance a topic, or a previous utterance in a
dialogue. These various conditions might be in conflict in the context
of a dialogue, and the model would need to rank (automatically) the 
conditions.

\section{Ethical Considerations}
\label{sec:ethical considerations}

\subsection{Models}
The proposed models are intended to link emotion theories from psychology and  
computational linguistics. The generated event descriptions can be used by psychologists
to study the impact of appraisal and emotions in written text. There are
several potential risks if the model is not used with care. It can result in 
biased or discriminatory language, despite that we have not observed
such behaviour. Potential reasons are that a model is trained on
biased data which could lead to generated texts that perpetuate
stereotypes or marginalize certain groups. Particularly in the case of
implicit expressions of emotions, it is important to employ models
with care.

In principle, models could be used for malicious purposes, for
instance to generate deceptive or harmful content (e.g., spreading
misinformation or generating fake news articles). Therefore, it is
crucial to employ responsible and ethical practices when utilizing
natural language generation models.  These risks are mainly inherent
from the base pre-train language models (\textit{Bart} and \textit{T5}) and they are not
intrinsic to our method.

\subsection{Human evaluation}
To conduct the human study in this research, we adhere to our
institutional regulations and follow the recommendations by the
Gemeinsame Ethikkomission der Hochschulen
Bayerns\footnote{\url{https://www.gehba.de/home/}} (GEHBa, Join Ethics
Committee of the Universities in Bavaria).  As per the guidelines
provided by the committee, studies that do not pose any specific risks
or burdens to participants beyond what they experience in their daily
lives do not require formal approval. Our study falls within that
category. Therefore, it did not require approval from an ethics
committee.

We relied on crowd-workers to conduct the human evaluation. The
annotators were recruited using \url{https://www.prolific.co}, 
and paid according to the platform rates (£9.00/hr).
All participants were shown a consent form containing the
information and requirements regarding the study. They
had to confirm their acceptance to be able to participate in 
the study. We provided an email address to contact us in case of problems during 
and after the study.

\section{Limitations}

Considering that our conditional approach is prompt-based, it is not
surprising that it has certain limitations. First, we mentioned that both Bart and T5 have
difficulties generating coherent and grammatical text, presumably because
of a limited compatibility between the conditional variables (\S
\ref{sec:experiments}). Second, the
conditions need to be represented as words or tokens and not numerical
representation (e.g., 1 or 0), since the models cannot identify the conditions
 and the prompt in the fine-tuning stage. Third, the
number of available datasets annotated with appraisals and emotions is
very limited, since the use of appraisal theories is relatively new in
the NLP community despite being a mature topic in psychology.

Even though appraisal conditions provided a better text generation for
a target emotion, through event descriptions, the text quality suffers
a small drop in quality (Table \ref{tab:human-grammar}). Overall, we
hope that the presented methodology and results can help guide future
research and rise interest in psychological appraisal theories.

 \section*{Acknowledgements}
 This work has been supported by a CONACYT scholarship
 (2020-000009-01EXTF-00195) and by the German
 Research Council (DFG), project
 ``Computational Event Analysis based on Appraisal Theories for Emotion
 Analysis'' (CEAT, project number KL 2869/1-2).

\bibliographystyle{acl_natbib}
\bibliography{lit}

\begin{thebibliography}{38}
\expandafter\ifx\csname natexlab\endcsname\relax\def\natexlab#1{#1}\fi

\bibitem[{Balahur et~al.(2011)Balahur, Hermida, Montoyo, and
  Mu{\~{n}}oz}]{Balahur2011}
Alexandra Balahur, Jes{\'u}s~M. Hermida, Andr{\'e}s Montoyo, and Rafael
  Mu{\~{n}}oz. 2011.
\newblock \href {https://doi.org/10.1007/978-3-642-22327-3_4} {Emotinet: A
  knowledge base for emotion detection in text built on the appraisal
  theories}.
\newblock In \emph{Natural Language Processing and Information Systems}, pages
  27--39, Berlin, Heidelberg. Springer Berlin Heidelberg.

\bibitem[{Brown et~al.(2020)Brown, Mann, Ryder, Subbiah, Kaplan, Dhariwal,
  Neelakantan, Shyam, Sastry, Askell, Agarwal, Herbert-Voss, Krueger, Henighan,
  Child, Ramesh, Ziegler, Wu, Winter, Hesse, Chen, Sigler, Litwin, Gray, Chess,
  Clark, Berner, McCandlish, Radford, Sutskever, and
  Amodei}]{NEURIPS2020_1457c0d6}
Tom Brown, Benjamin Mann, Nick Ryder, Melanie Subbiah, Jared~D Kaplan, Prafulla
  Dhariwal, Arvind Neelakantan, Pranav Shyam, Girish Sastry, Amanda Askell,
  Sandhini Agarwal, Ariel Herbert-Voss, Gretchen Krueger, Tom Henighan, Rewon
  Child, Aditya Ramesh, Daniel Ziegler, Jeffrey Wu, Clemens Winter, Chris
  Hesse, Mark Chen, Eric Sigler, Mateusz Litwin, Scott Gray, Benjamin Chess,
  Jack Clark, Christopher Berner, Sam McCandlish, Alec Radford, Ilya Sutskever,
  and Dario Amodei. 2020.
\newblock \href {https://proceedings.neurips.cc/paper/2020/file
  /1457c0d6bfcb4967418bfb8ac142f64a-Paper.pdf} {Language models are few-shot
  learners}.
\newblock In \emph{Advances in Neural Information Processing Systems},
  volume~33, pages 1877--1901. Curran Associates, Inc.

\bibitem[{Casel et~al.(2021)Casel, Heindl, and
  Klinger}]{casel-etal-2021-emotion}
Felix Casel, Amelie Heindl, and Roman Klinger. 2021.
\newblock \href {https://aclanthology.org/2021.konvens-1.5} {Emotion
  recognition under consideration of the emotion component process model}.
\newblock In \emph{Proceedings of the 17th Conference on Natural Language
  Processing (KONVENS 2021)}, pages 49--61, D{\"u}sseldorf, Germany. KONVENS
  2021 Organizers.

\bibitem[{Castro~Ferreira et~al.(2017)Castro~Ferreira, Calixto, Wubben, and
  Krahmer}]{ferreira2017linguistic}
Thiago Castro~Ferreira, Iacer Calixto, Sander Wubben, and Emiel Krahmer. 2017.
\newblock \href {https://doi.org/10.18653/v1/W17-3501} {Linguistic realisation
  as machine translation: Comparing different {MT} models for {AMR}-to-text
  generation}.
\newblock In \emph{Proceedings of the 10th International Conference on Natural
  Language Generation}, pages 1--10, Santiago de Compostela, Spain. Association
  for Computational Linguistics.

\bibitem[{Chan et~al.(2021)Chan, Ong, Pung, Zhang, and Fu}]{chan2020cocon}
Alvin Chan, Yew-Soon Ong, Bill Pung, Aston Zhang, and Jie Fu. 2021.
\newblock \href {https://openreview.net/forum?id=VD_ozqvBy4W} {Cocon: A
  self-supervised approach for controlled text generation}.
\newblock In \emph{International Conference on Learning Representations}.

\bibitem[{Cho et~al.(2014)Cho, van Merri{\"e}nboer, Gulcehre, Bahdanau,
  Bougares, Schwenk, and Bengio}]{cho-etal-2014-learning}
Kyunghyun Cho, Bart van Merri{\"e}nboer, Caglar Gulcehre, Dzmitry Bahdanau,
  Fethi Bougares, Holger Schwenk, and Yoshua Bengio. 2014.
\newblock \href {https://doi.org/10.3115/v1/D14-1179} {Learning phrase
  representations using {RNN} encoder{--}decoder for statistical machine
  translation}.
\newblock In \emph{Proceedings of the 2014 Conference on Empirical Methods in
  Natural Language Processing ({EMNLP})}, pages 1724--1734, Doha, Qatar.
  Association for Computational Linguistics.

\bibitem[{Colombo et~al.(2019)Colombo, Witon, Modi, Kennedy, and
  Kapadia}]{colombo-etal-2019-affect}
Pierre Colombo, Wojciech Witon, Ashutosh Modi, James Kennedy, and Mubbasir
  Kapadia. 2019.
\newblock \href {https://doi.org/10.18653/v1/N19-1374} {Affect-driven dialog
  generation}.
\newblock In \emph{Proceedings of the 2019 Conference of the North {A}merican
  Chapter of the Association for Computational Linguistics: Human Language
  Technologies, Volume 1 (Long and Short Papers)}, pages 3734--3743,
  Minneapolis, Minnesota. Association for Computational Linguistics.

\bibitem[{Cui et~al.(2022)Cui, Di, Shen, Ouchi, Liu, and Xu}]{cui2022modeling}
Fuwei Cui, Hui Di, Lei Shen, Kazushige Ouchi, Ze~Liu, and Jinan Xu. 2022.
\newblock \href {https://link.springer.com/article/10.1007/s10489-021-02683-x}
  {Modeling semantic and emotional relationship in multi-turn emotional
  conversations using multi-task learning}.
\newblock \emph{Applied Intelligence}, 52(4):4663--4673.

\bibitem[{Devlin et~al.(2019)Devlin, Chang, Lee, and
  Toutanova}]{devlin-etal-2019-bert}
Jacob Devlin, Ming-Wei Chang, Kenton Lee, and Kristina Toutanova. 2019.
\newblock \href {https://doi.org/10.18653/v1/N19-1423} {{BERT}: Pre-training of
  deep bidirectional transformers for language understanding}.
\newblock In \emph{"Proceedings of the 2019 Conference of the North {A}merican
  Chapter of the Association for Computational Linguistics: Human Language
  Technologies, Volume 1 (Long and Short Papers)"}, pages 4171--4186,
  Minneapolis, Minnesota. Association for Computational Linguistics.

\bibitem[{Ekman(1992)}]{ekman1992argument}
Paul Ekman. 1992.
\newblock \href
  {https://www.paulekman.com/wp-content/uploads/2013/07/An-Argument-For-Basic
  -Emotions.pdf} {An argument for basic emotions}.
\newblock \emph{Cognition \& emotion}, 6(3-4):169--200.

\bibitem[{Ekman and Davidson(1994)}]{ekman1994nature}
Paul Ekman and Richard~J Davidson. 1994.
\newblock \href {https://doi.apa.org/record/1995-97541-000} {\emph{The nature
  of emotion: Fundamental questions.}}
\newblock Oxford University Press.

\bibitem[{Ellsworth and Smith(1988)}]{Ellsworth1988}
Phoebe~C. Ellsworth and Craig~A. Smith. 1988.
\newblock From appraisal to emotion: Differences among unpleasant feelings.
\newblock \emph{Motivation and emotion}, 12(3):271--302.

\bibitem[{Ghosh et~al.(2017)Ghosh, Chollet, Laksana, Morency, and
  Scherer}]{ghosh-etal-2017-affect}
Sayan Ghosh, Mathieu Chollet, Eugene Laksana, Louis-Philippe Morency, and
  Stefan Scherer. 2017.
\newblock \href {https://doi.org/10.18653/v1/P17-1059} {Affect-{LM}: A neural
  language model for customizable affective text generation}.
\newblock In \emph{Proceedings of the 55th Annual Meeting of the Association
  for Computational Linguistics (Volume 1: Long Papers)}, pages 634--642,
  Vancouver, Canada. Association for Computational Linguistics.

\bibitem[{Green(1992)}]{Green1992}
Otis~H. Green. 1992.
\newblock \href {https://doi.org/10.1007/978-94-011-2552-9_6} {\emph{The
  Belief-Desire Theory of Emotions}}, pages 77--106. Springer Netherlands,
  Dordrecht.

\bibitem[{Hofmann et~al.(2020)Hofmann, Troiano, Sassenberg, and
  Klinger}]{Hofmann2020b}
Jan Hofmann, Enrica Troiano, Kai Sassenberg, and Roman Klinger. 2020.
\newblock \href {https://doi.org/10.18653/v1/2020.coling-main.11} {Appraisal
  theories for emotion classification in text}.
\newblock In \emph{Proceedings of the 28th International Conference on
  Computational Linguistics}, pages 125--138, Barcelona, Spain (Online).
  International Committee on Computational Linguistics.

\bibitem[{Kondadadi et~al.(2013)Kondadadi, Howald, and
  Schilder}]{kondadadi2013statistical}
Ravi Kondadadi, Blake Howald, and Frank Schilder. 2013.
\newblock \href {https://aclanthology.org/P13-1138} {A statistical {NLG}
  framework for aggregated planning and realization}.
\newblock In \emph{Proceedings of the 51st Annual Meeting of the Association
  for Computational Linguistics (Volume 1: Long Papers)}, pages 1406--1415,
  Sofia, Bulgaria. Association for Computational Linguistics.

\bibitem[{Lebret et~al.(2016)Lebret, Grangier, and Auli}]{lebret2016neural}
R{\'e}mi Lebret, David Grangier, and Michael Auli. 2016.
\newblock \href {https://doi.org/10.18653/v1/D16-1128} {Neural text generation
  from structured data with application to the biography domain}.
\newblock In \emph{Proceedings of the 2016 Conference on Empirical Methods in
  Natural Language Processing}, pages 1203--1213, Austin, Texas. Association
  for Computational Linguistics.

\bibitem[{Lewis et~al.(2020)Lewis, Liu, Goyal, Ghazvininejad, Mohamed, Levy,
  Stoyanov, and Zettlemoyer}]{lewis2019bart}
Mike Lewis, Yinhan Liu, Naman Goyal, Marjan Ghazvininejad, Abdelrahman Mohamed,
  Omer Levy, Veselin Stoyanov, and Luke Zettlemoyer. 2020.
\newblock \href {https://doi.org/10.18653/v1/2020.acl-main.703} {{BART}:
  Denoising sequence-to-sequence pre-training for natural language generation,
  translation, and comprehension}.
\newblock In \emph{Proceedings of the 58th Annual Meeting of the Association
  for Computational Linguistics}, pages 7871--7880, Online. Association for
  Computational Linguistics.

\bibitem[{Liu et~al.(2019)Liu, Ott, Goyal, Du, Joshi, Chen, Levy, Lewis,
  Zettlemoyer, and Stoyanov}]{liu2019roberta}
Yinhan Liu, Myle Ott, Naman Goyal, Jingfei Du, Mandar Joshi, Danqi Chen, Omer
  Levy, Mike Lewis, Luke Zettlemoyer, and Veselin Stoyanov. 2019.
\newblock \href {https://arxiv.org/abs/1907.11692} {Roberta: A robustly
  optimized bert pretraining approach}.
\newblock \emph{arXiv preprint arXiv:1907.11692}.

\bibitem[{Lowerre(1976)}]{lowerre1976harpy}
Bruce~T Lowerre. 1976.
\newblock \emph{The harpy speech recognition system.}
\newblock Carnegie Mellon University.

\bibitem[{Maqsud(2015)}]{maqsud2015synthetic}
Umar Maqsud. 2015.
\newblock \href {https://doi.org/10.18653/v1/W15-2922} {Synthetic text
  generation for sentiment analysis}.
\newblock In \emph{Proceedings of the 6th Workshop on Computational Approaches
  to Subjectivity, Sentiment and Social Media Analysis}, pages 156--161,
  Lisboa, Portugal. Association for Computational Linguistics.

\bibitem[{Mohammad(2012)}]{Mohammad12}
Saif Mohammad. 2012.
\newblock \href {https://www.aclweb.org/anthology/S12-1033} {{\#}emotional
  tweets}.
\newblock In \emph{*{SEM} 2012: The First Joint Conference on Lexical and
  Computational Semantics {--} Volume 1: Proceedings of the main conference and
  the shared task, and Volume 2: Proceedings of the Sixth International
  Workshop on Semantic Evaluation ({S}em{E}val 2012)}, pages 246--255,
  Montr{\'e}al, Canada. Association for Computational Linguistics.

\bibitem[{Niu and Bansal(2018)}]{niu2018polite}
Tong Niu and Mohit Bansal. 2018.
\newblock \href {https://transacl.org/ojs/index.php/tacl/article/view/1424}
  {Polite dialogue generation without parallel data}.
\newblock \emph{Transactions of the Association for Computational Linguistics},
  6:373--389.

\bibitem[{Orbach and Goldberg(2020)}]{orbach2020facts2story}
Eyal Orbach and Yoav Goldberg. 2020.
\newblock \href {https://doi.org/10.18653/v1/2020.coling-main.211}
  {{F}acts2{S}tory: Controlling text generation by key facts}.
\newblock In \emph{Proceedings of the 28th International Conference on
  Computational Linguistics}, pages 2329--2345, Barcelona, Spain (Online).
  International Committee on Computational Linguistics.

\bibitem[{Plutchik(2001)}]{plutchik2001nature}
Robert Plutchik. 2001.
\newblock \href {https://www.jstor.org/stable/27857503?seq=1} {The nature of
  emotions: Human emotions have deep evolutionary roots, a fact that may
  explain their complexity and provide tools for clinical practice}.
\newblock \emph{American scientist}, 89(4):344--350.

\bibitem[{Plutchik and Kellerman(2013)}]{plutchik2013theories}
Robert Plutchik and Henry Kellerman. 2013.
\newblock \emph{Theories of emotion}, volume~1.
\newblock Academic Press.

\bibitem[{Radford et~al.(2018)Radford, Narasimhan, Salimans, and
  Sutskever}]{radford2018improving}
Alec Radford, Karthik Narasimhan, Tim Salimans, and Ilya Sutskever. 2018.
\newblock \href {https://www.gwern.net/docs/www/s3-us-west-2.amazonaws.com
  /d73fdc5ffa8627bce44dcda2fc012da638ffb158.pdf} {Improving language
  understanding by generative pre-training}.
\newblock \emph{OpenAI blog}.

\bibitem[{Radford et~al.(2019)Radford, Wu, Child, Luan, Amodei, Sutskever
  et~al.}]{radford2019language}
Alec Radford, Jeffrey Wu, Rewon Child, David Luan, Dario Amodei, Ilya
  Sutskever, et~al. 2019.
\newblock \href
  {https://d4mucfpksywv.cloudfront.net/better-language-models/language-models.pdf}
  {Language models are unsupervised multitask learners}.
\newblock \emph{OpenAI blog}, 1(8):9.

\bibitem[{Raffel et~al.(2020)Raffel, Shazeer, Roberts, Lee, Narang, Matena,
  Zhou, Li, and Liu}]{raffel2019exploring}
Colin Raffel, Noam Shazeer, Adam Roberts, Katherine Lee, Sharan Narang, Michael
  Matena, Yanqi Zhou, Wei Li, and Peter~J. Liu. 2020.
\newblock \href {http://jmlr.org/papers/v21/20-074.html} {Exploring the limits
  of transfer learning with a unified text-to-text transformer}.
\newblock \emph{Journal of Machine Learning Research}, 21(140):1--67.

\bibitem[{Scherer et~al.(2001)Scherer, Schorr, and
  Johnstone}]{scherer2001appraisal}
Klaus~R Scherer, Angela Schorr, and Tom Johnstone. 2001.
\newblock \emph{Appraisal processes in emotion: Theory, methods, research}.
\newblock Oxford University Press.

\bibitem[{Smith and Ellsworth(1985)}]{Smith1985}
Craig.~A. Smith and Phoebe.~C. Ellsworth. 1985.
\newblock Patterns of cognitive appraisal in emotion.
\newblock \emph{Journal of Personality and Social Psychology}, 48(4):813--838.

\bibitem[{Song et~al.(2019)Song, Zheng, Liu, Xu, and
  Huang}]{song2019generating}
Zhenqiao Song, Xiaoqing Zheng, Lu~Liu, Mu~Xu, and Xuanjing Huang. 2019.
\newblock \href {https://doi.org/10.18653/v1/P19-1359} {Generating responses
  with a specific emotion in dialog}.
\newblock In \emph{Proceedings of the 57th Annual Meeting of the Association
  for Computational Linguistics}, pages 3685--3695, Florence, Italy.
  Association for Computational Linguistics.

\bibitem[{Sutskever et~al.(2014)Sutskever, Vinyals, and
  Le}]{sutskever2014sequence}
Ilya Sutskever, Oriol Vinyals, and Quoc~V Le. 2014.
\newblock \href {https://proceedings.neurips.cc/paper/2014/file
  /a14ac55a4f27472c5d894ec1c3c743d2-Paper.pdf} {Sequence to sequence learning
  with neural networks}.
\newblock \emph{Advances in neural information processing systems}, 27.

\bibitem[{Troiano et~al.(2022)Troiano, Oberlaender, Wegge, and
  Klinger}]{Troiano2022-envent}
Enrica Troiano, Laura Ana~Maria Oberlaender, Maximilian Wegge, and Roman
  Klinger. 2022.
\newblock \href {https://aclanthology.org/2022.lrec-1.146} {x-envent: A corpus
  of event descriptions with experiencer-specific emotion and appraisal
  annotations}.
\newblock In \emph{Proceedings of the Language Resources and Evaluation
  Conference}, pages 1365--1375, Marseille, France. European Language Resources
  Association.

\bibitem[{Troiano et~al.(2023)Troiano, Oberl\"ander, and Klinger}]{Troiano2023}
Enrica Troiano, Laura Oberl\"ander, and Roman Klinger. 2023.
\newblock \href {https://doi.org/10.1162/coli_a_00461} {Dimensional modeling of
  emotions in text with appraisal theories: Corpus creation, annotation
  reliability, and prediction}.
\newblock \emph{Computational Linguistics}, 49(1).

\bibitem[{Troiano et~al.(2019)Troiano, Pad{\'o}, and
  Klinger}]{troiano-etal-2019-crowdsourcing}
Enrica Troiano, Sebastian Pad{\'o}, and Roman Klinger. 2019.
\newblock \href {https://doi.org/10.18653/v1/P19-1391} {Crowdsourcing and
  validating event-focused emotion corpora for {G}erman and {E}nglish}.
\newblock In \emph{Proceedings of the 57th Annual Meeting of the Association
  for Computational Linguistics}, pages 4005--4011, Florence, Italy.
  Association for Computational Linguistics.

\bibitem[{Zhang et~al.(2019)Zhang, Wang, Yin, and Huang}]{zhang2019emotional}
Rui Zhang, Zhenyu Wang, Kai Yin, and Zhenhua Huang. 2019.
\newblock \href {https://doi.org/10.1109/ACCESS.2019.2931036} {Emotional text
  generation based on cross-domain sentiment transfer}.
\newblock \emph{IEEE Access}, 7:100081--100089.

\bibitem[{Zhou et~al.(2018)Zhou, Huang, Zhang, Zhu, and
  Liu}]{zhou2018emotional}
Hao Zhou, Minlie Huang, Tianyang Zhang, Xiaoyan Zhu, and Bing Liu. 2018.
\newblock \href {https://doi.org/10.5555/3504035.3504125} {Emotional chatting
  machine: Emotional conversation generation with internal and external
  memory}.
\newblock In \emph{Proceedings of the Thirty-Second AAAI Conference on
  Artificial Intelligence and Thirtieth Innovative Applications of Artificial
  Intelligence Conference and Eighth AAAI Symposium on Educational Advances in
  Artificial Intelligence}, AAAI'18/IAAI'18/EAAI'18, page 730–738. AAAI
  Press.

\end{thebibliography}

\appendix
\clearpage

\section{Filtered Crowd-enVent Dataset}
\label{sec:filtered_enVent}

As described in \S\ref{sub:experiments},  we examine seven emotions
(Anger, Disgust, Fear, Guilt,
Joy, Sadness, and Shame), and seven appraisals (Attention, Responsibility,
Control, Circumstance, Pleasantness, Effort, and
Circumstance) as conditional variables. Therefore, we filter the crowd-enVent dataset by
removing records that do not have one of the seven emotions with at
least one of the seven emotions. We follow the same criteria proposed
by \citet{Troiano2023} to discretize the emotion and appraisal values
(1 if the annotator score is larger than 3, else 0). Table
\ref{tab:envent-stats} provides the statistical analysis of the
filtered dataset. It shows the co-occurrence between emotions
and appraisals, as well as details about the text, including the 
number of tokens, verbs, adjectives, nouns, and clauses.

\section{Automatic Classifiers}
\label{sec:Auto Classifiers}

\begin{table}
  \centering\small
  \setlength{\tabcolsep}{8pt}
  \renewcommand{\arraystretch}{0.8}
  \begin{tabularx}{\linewidth}{Xrrr}
    \toprule
    Appraisal & Precision & Recall & F$_1$ \\
    \cmidrule(r){1-1}\cmidrule(rl){2-2}\cmidrule(l){3-3}\cmidrule(l){4-4}
    Attention & .68 & .66 & .66 \\
    Certainty & .51 & .39 & .38 \\
    Circumstance & .60 & .57 & .58 \\
    Control & .56 & .56 & .56 \\
    Effort & .54 & .53 & .52 \\
    Pleasantness & .63 & .59 & .60 \\
    Responsibility & .60 & .58 & .59 \\
    \cmidrule(r){1-1}\cmidrule(rl){2-2}\cmidrule(l){3-3}\cmidrule(l){4-4}
    Macro-Avg. & .59 & .55 & .56 \\ \bottomrule
  \end{tabularx}
  \caption{Precision, Recall and F$_1$ scores from the appraisal
    classifiers.}
  \label{tab:app-scores}
\end{table}

To get an impression of the reliability of the different model
architectures (\textit{Bart} and \textit{T5}) with different
conditional configurations (EA, E, A), we train one multi-label
classifier for the seven emotions and 7 binary classifiers for each
appraisal. The classifiers are built on top of RoBERTa
\citep{liu2019roberta} using the standard parameters for ten epochs
with a batch size of five. Please refer to Table~\ref{tab:app-scores}
for precision, recall, and F$_1$ scores of the appraisal classifiers,
and Table~\ref{tab:emo_scores} for the corresponding scores related to
emotions.

The results for automatic classification of the appraisals are
presented in Table~\ref{tab:cnlg-app}.  We observed that appraisal
information improves the performance for emotion accuracy. This cannot
be observed for the appraisal variables. For most appraisal
dimensions, the model that is not conditioned on emotions works better
(A is better than EA). The gap between EA and E for the same
architecture is 7\,pp for \textit{T5}, and 1\,pp for \textit{Bart}.

\begin{table}[t]
 \centering\small
 \setlength{\tabcolsep}{4pt}
\renewcommand{\arraystretch}{0.9}
\begin{tabular}{lllrrrrrrrr}
\toprule
\rt{Arch.} & \rt{Conf.} & \rt{\shortstack{Testing\\ Prmpt.}} & \rt{Att.} & \rt{Resp.} & \rt{Contr.} &
\rt{Circ.} & \rt{Plea.} & \rt{Effo.} & \rt{Cert.} & \rt{M. Avg.}\\
\cmidrule(r){1-1}\cmidrule(rl){2-2}\cmidrule(rl){3-3}\cmidrule(rl){4-4
}\cmidrule(rl){5-5}\cmidrule(rl){6-6}\cmidrule(rl){7-7}\cmidrule(rl){8-8}
\cmidrule(rl){9-9}\cmidrule(rl){10-10}\cmidrule(l){11-11}
T5 & EA & AP & \textbf{.45} & .42 & .36 & .52 & .50 & .47 & .37 & .44 \\
T5 & A & AP & \textbf{.45} & \textbf{.48} & .44 & \textbf{.71} & \textbf{.66} &
.46 & \textbf{.38} & \textbf{.51} \\

\cmidrule(r){1-1}\cmidrule(rl){2-2}\cmidrule(rl){3-3}\cmidrule(rl){4-4
}\cmidrule(rl){5-5}\cmidrule(rl){6-6}\cmidrule(rl){7-7}\cmidrule(rl){8-8}
\cmidrule(l){9-9}\cmidrule(rl){10-10}\cmidrule(l){11-11}
 
Bart & EA & AP & \textbf{.45} & .43 & .42 & .53 & .47 & \textbf{.50} & .35 &
.45
\\
Bart & A & AP & .35 & .43 & \textbf{.50} & .57 & .60 & .48 & .35 & .46 \\

\bottomrule
\end{tabular}
\caption{Appraisal F$_1$ score over the generated text using the AP Prompt set,
from the models conditioned on emotion and appraisals (EA), and appraisals
(A).}
\label{tab:cnlg-app}
\end{table}

\begin{table}
  \centering\small
  \setlength{\tabcolsep}{8pt}
  \renewcommand{\arraystretch}{0.8}
  \begin{tabularx}{\linewidth}{Xrrr}
    \toprule
    Emotion & Precision & Recall & F$_1$ \\ 
    \cmidrule(r){1-1}\cmidrule(rl){2-2}\cmidrule(l){3-3}\cmidrule(l){4-4}
    Anger & .72 & .58 & .64 \\
    Disgust & .74 & .80 & .77 \\
    Fear & .78 & .93 & .85 \\
    Guilt & .56 & .71 & .62 \\
    Joy & .91 & .92 & .98 \\
    Sadness & .91 & .87 & .89 \\
    Shame & .66 & .43 & .52 \\
    \cmidrule(r){1-1}\cmidrule(rl){2-2}\cmidrule(l){3-3}\cmidrule(l){4-4}
    Macro-Avg. & .75 & .75 & .75 \\ \bottomrule
  \end{tabularx}
  \caption{Precision, Recall and F$_1$ scores from the emotion
    classifier over the 7 classes.}
  \label{tab:emo_scores}
\end{table}

\section{Human Evaluation Study Details}
\subsection{Study Details}
\label{sec:Human_suervey}

The human evaluation is performed on 330 sentences, 30 human-generated
sentences from the crow-enVent dataset, and 100 sentences randomly
selected from each of the following model configurations and prompt
sets: EA with EP, E with EP, and EA with EfA. We use human-generated
sentences to validate the study as a gold standard, under the
assumption that humans are capable of accurately evaluating text
written by other humans. For this purpose, we selected the top 30
\textit{easy} sentences by ranking the filtered crowd-enVent dataset
using two metrics: Emotion agreement and appraisal agreement. Table
\ref{tab:prolific-stats} shows the statistical analysis of the 330
sentences.

\begin{table}
  \centering\small
\setlength{\tabcolsep}{2.5pt}
\renewcommand{\arraystretch}{0.8}
\begin{tabular}{llrrrrr}
\toprule
\rt{Conf.} &  \rt{\shortstack{Testing\\ Prmpt.}}& \rt{\shortstack{Tokens\\ (std.)}} &
\rt{\shortstack{Nouns\\ (std.)}} & \rt{\shortstack{Verbs\\ (std.)}} &
\rt{\shortstack{Adj.\\ (std.)}} & \rt{\shortstack{Clauses\\ (std.)}} \\
\cmidrule(r){1-1}\cmidrule(rl){2-2}\cmidrule(rl){3-3}\cmidrule(rl){4-4
}\cmidrule(rl){5-5}\cmidrule(rl){6-6}\cmidrule(l){7-7}

Hum. & enVent & \countstd{22.8}{16.8} & \countstd{4.4}{3.2} &
\countstd{3.3}{2.4} & \countstd{1.2}{1.8} & \countstd{1.7}{.7} \\

\cmidrule(r){1-1}\cmidrule(rl){2-2}\cmidrule(rl){3-3}\cmidrule(rl){4-4
}\cmidrule(rl){5-5}\cmidrule(rl){6-6}\cmidrule(l){7-7}

EA & EP & \countstd{15.3}{4.0} & \countstd{2.4}{1.0} & \countstd{2.2}{1.0} &
\countstd{.7}{.8} & \countstd{1.5}{.6}\\

EA & EfA & \countstd{13.7}{4.7} & \countstd{1.8}{1.2} & \countstd{2.1}{1.2} &
\countstd{.6}{.9} & \countstd{1.4}{.6}\\

E & EP & \countstd{9.2}{3.6} & \countstd{1.6}{1.0} & \countstd{1.6}{0.8} &
\countstd{.5}{.7} & \countstd{1.3}{.5}\\

\bottomrule
\end{tabular}
\caption{Statistical analysis of the automatically and human-generated text
for human evaluation.}
\label{tab:prolific-stats}
\end{table}

\begin{table*}
  \centering\small
\setlength{\tabcolsep}{5pt}
\renewcommand{\arraystretch}{0.8}
\begin{tabular}{lrrrrrrrrrrrrr}
\toprule

\rt{Emo} &  \rt{Docs.}& \rt{Att.} & \rt{Resp.} & \rt{Contr.} & \rt{Circ.} &
\rt{Plea.} & \rt{Effo.} & \rt{Cert.} & \rt{\shortstack{Tokens\\ (std.)}} &
\rt{\shortstack{Nouns\\ (std.)}} & \rt{\shortstack{Verbs\\ (std.)}} &
\rt{\shortstack{Adj.\\ (std.)}} & \rt{\shortstack{Clauses\\ (std.)}} \\

\cmidrule(r){1-1} \cmidrule(rl){2-2} \cmidrule(rl){3-3} \cmidrule(rl){4-4}
\cmidrule(rl){5-5} \cmidrule(rl){6-6} \cmidrule(rl){7-7} \cmidrule(rl){8-8}
\cmidrule(rl){9-9} \cmidrule(rl){10-10} \cmidrule(rl){11-11}
\cmidrule(rl){12-12} \cmidrule(rl){13-13} \cmidrule(rl){13-13}
\cmidrule(l){14-14}

Ang. & 450 & 305 & 55 & 86 & 72 & 15 & 309 & 184 & 21.8 \std{30.8} & 3.7
\std{4.4} & 3.2 \std{4.4} &  0.9 \std{1.8} & 1.4 \std{0.7}  \\

Dis. & 450 & 228 & 66 & 90 & 103 & 6 & 193 & 155 & 19.4 \std{19.1} & 3.7
 \std{3.4} & 2.8 \std{2.8} & 1.0 \std{1.5}  & 1.4 \std{0.6} \\

Fear. & 450 & 378 & 119 & 100 & 157 & 17 & 345 & 148 & 19.4 \std{24.5} & 3.4
\std{3.9} & 2.8 \std{3.7} & 1.0 \std{1.4} & 1.3 \std{0.7} \\

Guilt. & 225 & 129 & 168 & 119 & 33 & 16 & 119 & 109 & 20.5 \std{22.1} & 3.2
\std{2.9} & 3.13 \std{3.4} & 1.0 \std{1.5} & 1.3  \std{0.6}\\

Joy. & 450 & 292 & 274 & 240 & 77 & 417 & 192 & 241 & 17.9 \std{20.7} & 3.2
\std{3.2} & 2.5 \std{2.9} & 1.1 \std{1.5} & 1.2 \std{0.5}\\

Sad. & 450 & 290 & 94 & 65 & 200 & 5 & 336 & 189 & 18.9 \std{22.8} & 2.9
\std{3.3} & 2.9 \std{3.4} & 1.0 \std{1.6} & 1.3 \std{0.6}\\

Shame. & 225 & 140 & 163 & 93 & 37 & 9 & 125 & 100 & 18.4 \std{22.4} & 2.8
\std{3.1} & 2.9 \std{3.6} & 0.8 \std{1.2} & 1.4 \std{0.7} \\

\cmidrule(r){1-1} \cmidrule(rl){2-2} \cmidrule(rl){3-3} \cmidrule(rl){4-4}
\cmidrule(rl){5-5} \cmidrule(rl){6-6} \cmidrule(rl){7-7} \cmidrule(rl){8-8}
\cmidrule(rl){9-9} \cmidrule(rl){10-10} \cmidrule(rl){11-11}
\cmidrule(rl){12-12} \cmidrule(rl){13-13} \cmidrule(rl){13-13}
\cmidrule(l){14-14} 

Total/Avg. &  2700 & 1762 & 939 & 793 & 679 & 485 & 1619 & 1126 & \countstd{19.5}{23.7} 
&\countstd{3.3}{3.7}& \countstd{2.9}{3.5} & \countstd{1.0}{1.5}
& \countstd{1.4}{0.6}\\

\bottomrule
\end{tabular}
\caption{Statistical analysis of the filtered crowd en-Vent dataset.
  Appraisal columns show the co-occurrence of a given appraisal and
  one emotion (row). Token, Nouns, Adj., and Clauses columns are the
  average counts for each instance.}
\label{tab:envent-stats}
\end{table*}

The survey was deployed on \url{https://www.soscisurvey.de}, and it
consists of 23 questions (Table \ref{tab:human-survey}), divided into
three sections of seven statements each, and two attention checks in a
random position. The first section evaluates the emotion category of the
text, the second the appraisal perception, and the last one, the
quality of the text. We ask the annotator how much they agree to each
statement using a five-level Likert scale (Not at all, Slightly,
Somewhat, Moderately, and Extremely).

The study was conducted in August 2022, at a total cost of £250.74.
Each text was annotated by three different annotators. The
annotators were recruited using \url{https://www.prolific.co}
with the following criteria:
\begin{compactitem}
\item Age: Minimum 18 and Maximum 50.
\item Nationality: UK, USA, IE.
\item Place of most time spent before turning 18: United Kingdom, United
  States, Ireland.
\item First language: English.
\item Approval rate: Minimum approval rate .75.
\end{compactitem}

\begin{table}
  \centering\small
\setlength{\tabcolsep}{3pt}
\begin{tabular}{lllrrrrrrrr}
\toprule

& \rt{Conf.} &  \rt{\shortstack{Testing\\ Prmpt.}} & \rt{Att.} & \rt{Resp.} & \rt{Contr.}
& \rt{Circ.} & \rt{Plea.} & \rt{Effo.} & \rt{Cert.} & \rt{M. Avg.} \\

\cmidrule(r){2-2}\cmidrule(rl){3-3}\cmidrule(rl){4-4
}\cmidrule(rl){5-5}\cmidrule(rl){6-6}\cmidrule(rl){7-7}\cmidrule(rl){8-8
}\cmidrule(rl){9-9}\cmidrule(r){10-10}\cmidrule(rl){11-11}

\multirow{2}{*}{\rt{Hum.}}
& Hum. & enVent & .94 & .88 & .69 & .71 & .85 & .77 & .60
& .78\\
& EA & EfA & .72 & .63 & .54 & .37 & .6 & .67 & .55 & .58\\

\cmidrule(rl){1-11}
\multirow{2}{*}{\rt{Auto.}}
& Hum. & enVent & .71 & .74 & .53 & .64 & .92 & .38 & .48 & .63\\
& EA & EfA & .57 & .63 & .5 & .36 & .24 & .12 &
.49 & .42\\
\bottomrule
\end{tabular}
\caption{Human annotation results as \F (1st and 2nd row) and
  automatic classification results (3rd and 4th row) of the human
  generated text (1st and 3rd row) and the automatically generated
  text (2nd, and 4th).}
\label{tab:human-appraisals}
\end{table}

\subsection{Appraisal Results}
\label{sec:Human_eval_Appraisals}
In the human evaluation in \S\ref{sec:Human-eva}, we mainly focus on
emotion evaluation. We now discuss briefly the results regarding
appraisal variables.

The appraisal evaluation (Table \ref{tab:human-appraisals}) exhibits similar
behavior to \S\ref{sec:Qua-Analysis}; the results for both
automatic and human evaluation are similar (2nd and 4th row). Therefore, it can
be inferred that state-of-the-art classifiers are as good as humans, and that
appraisal classification is a hard task. Even with \textit{easy} texts (1st
row) humans only achieve  $78\,\%$ (while for emotions they achieve
$100\,\%$). These results are aligned with \citet{Troiano2022-envent}.

\begin{table}
  \centering\small
\setlength{\tabcolsep}{2pt}
\renewcommand{\arraystretch}{0.6}
\begin{tabular}{cp{0.4\textwidth}}
\toprule

Sec. & Statements \\

\cmidrule(r){1-1}\cmidrule(l){2-2}

  \multirow{12}{*}{\rt{Appraisal}}
        & \textbf{How much do these statements apply?} \\\cmidrule{2-2}
        & The experiencer had to pay attention to the situation. \\
        & The event was caused by the experiencer’s own behavior.\\
& The experiencer was able to influence what was going on during the event.\\
& The situation was the result of outside influences over which nobody had
          control.\\
        & The event was pleasant for the experiencer.\\
        & The situation required her/him a great deal of energy.\\
        & The experiencer anticipated the consequence of the event.\\

\cmidrule{1-2}
\multirow{10}{*}{\rt{Emotion}}
        & \textbf{What do you think the writer of the text felt when
          experiencing this event?} \\\cmidrule{2-2}
        & Anger.\\
        & Disgust.\\
        & Fear.\\
        & Guilt.\\
        & Joy.\\
        & Sadness.\\
        & Shame.\\

\cmidrule{1-2}
\multirow{8}{*}{\rt{Text quality}}
        & \textbf{How understandable is
          the text for you?} \\\cmidrule{2-2}
        & The text is fluent. \\
        & The text has grammatical issues.\\
        & The text is written by a native English speaker.\\
        & The text is semantically coherent.\\
        & What the text describes might have really happened.\\
        & The text has been written by an artificial intelligence/machine.\\
        & The text has been written by a human.\\

\cmidrule{1-2}
\multirow{3}{*}{\rt{A.C.}}
& Attention check. Please click “Moderately”. \\
& The current question is an attention check, please select “Extremely”.\\

\bottomrule
\end{tabular}
\caption{Human evaluation survey}
\label{tab:human-survey}
\end{table}

\end{document}